\documentclass{article}
\usepackage{ijcai24}

\usepackage{times}
\usepackage{soul}
\usepackage{url}
\usepackage[hidelinks]{hyperref}
\usepackage[utf8]{inputenc}
\usepackage[small]{caption}
\usepackage{graphicx}
\usepackage{amsmath}
\usepackage{amsthm}
\usepackage{booktabs}
\usepackage{algorithm}
\usepackage{algorithmic}
\usepackage[switch]{lineno}

\usepackage{algorithm}
\usepackage{algorithmic}
\usepackage{graphicx}
\usepackage[mathscr]{eucal}
\usepackage{amsmath}
\usepackage{amssymb}
\usepackage[table,xcdraw]{xcolor}
\usepackage{multirow}

\usepackage{bm}
\usepackage{xcolor}

\title{PoCo: A Self-Supervised Approach via Polar Transformation Based Progressive Contrastive Learning for Ophthalmic Disease Diagnosis}

\author{
Jinhong Wang$^1$
\and
Tingting Chen$^2$\and
Jintai Chen$^3$\and
Danny Chen$^4$\and
Haochao Ying$^1$\and
Jian Wu$^1$\\
\affiliations
$^1$Zhejiang University \quad $^2$University of Pennsylvania\\
$^3$ University of Illinois Urbana-Champaign \quad
$^4$University of Notre Dame\\
\emails
wangjinhong@zju.edu.cn,
ttchen0603@gmail.com,
jtchen721@gmail.com, 
dchen@nd.edu, 
\{haochaoying, wujian2000\}@zju.edu.cn
}

\begin{document}

\maketitle

\begin{abstract}
    Automatic ophthalmic disease diagnosis on fundus images is important in clinical practice. However, due to complex fundus textures and limited annotated data, developing an effective automatic method for this problem is still challenging. In this paper, we present a self-supervised method via {\underline{Po}}lar transformation-based progressive {\underline{Co}}ntrastive learning, called PoCo, for ophthalmic disease diagnosis. Specifically, we novelly inject the polar transformation into contrastive learning to 1) promote contrastive learning pre-training to be faster and more stable and 2) naturally capture task-free and rotation-related textures, which provides insights into disease recognition on fundus images. Beneficially, simple normal translation-invariant convolution on transformed images can equivalently replace the complex rotation-invariant and sector convolution on raw images. After that, we develop a progressive contrastive learning method to efficiently utilize large unannotated images and a novel progressive hard negative sampling scheme to gradually reduce the negative sample number for efficient training and performance enhancement. Extensive experiments on three public ophthalmic disease datasets show that our PoCo achieves state-of-the-art performance with good generalization ability, validating that our method can reduce annotation efforts and provide reliable diagnosis. Codes link:  \url{https://github.com/wjh892521292/PoCo}.
\end{abstract}

\section{Introduction}

In clinical practice, fundus images are often used to diagnose various ophthalmic diseases, including glaucoma, diabetic retinopathy (DR)~\cite{he2020cabnet,li2019canet}, age-related macular degeneration (AMD)~\cite{age2000risk,yim2020predicting}, cataract~\cite{zhang2019automatic}, pathological myopia (PM)~\cite{morgan2012myopia}, diabetic macular edema (DME)~\cite{sahlsten2019deep}, and more. Recently, automatic computer-aided methods have been applied to ophthalmic disease diagnosis with fundus images based on deep learning (DL)~\cite{chen2015glaucoma,he2020cabnet,li2019canet,sahlsten2019deep,yim2020predicting,zhang2019automatic}. However, these DL methods commonly require a large amount of labeled data for model training. 
Unfortunately, acquiring labeled data is highly expensive due to the tedious and laborious annotation process, even for experienced doctors. Self-supervised learning (SSL), an advanced and generic representation learning paradigm, can be used to tackle this challenge efficiently by first pre-training with unlabeled data and then fine-tuning for a downstream task with limited labeled data. 
Therefore, exploring self-supervised learning to reduce the labeling cost of fundus images for ophthalmic disease diagnosis is of great importance.

In the medical imaging domain, due to the potential of training with a large amount of unlabeled data, SSL has been widely used in various types of disease diagnosis (e.g., cardiac MR image segmentation~\cite{bai2019self}, nodule detection~\cite{tajbakhsh2019surrogate}, and brain hemorrhage classification~\cite{zhuang2019self}). For 
ophthalmic disease diagnosis, as contrastive learning (CL, a type of SSL) in recent years showed powerful representation capabilities in self-supervised unlabeled training, more CL methods were developed. Li et al.~\cite{li2020self} applied CL to multi-modal fundus images for retinal disease diagnosis. Li et al.~\cite{li2021rotation} devised a rotation-oriented collaborative method including a rotation prediction task and a multi-instance discrimination CL task for retinal disease diagnosis. But, these methods did not consider the specificity of ophthalmic disease analysis on fundus images. For example, the features used for diagnosis are usually annular (e.g., vessel structures), which are different from natural images. How to better extract such annular texture features is a key problem. 
Worse, known CL methods also neglect negative sample selection (especially hard samples), which could hinder model performance considerably.

\begin{figure*}[t]
\centering
\includegraphics[width=0.89\textwidth]{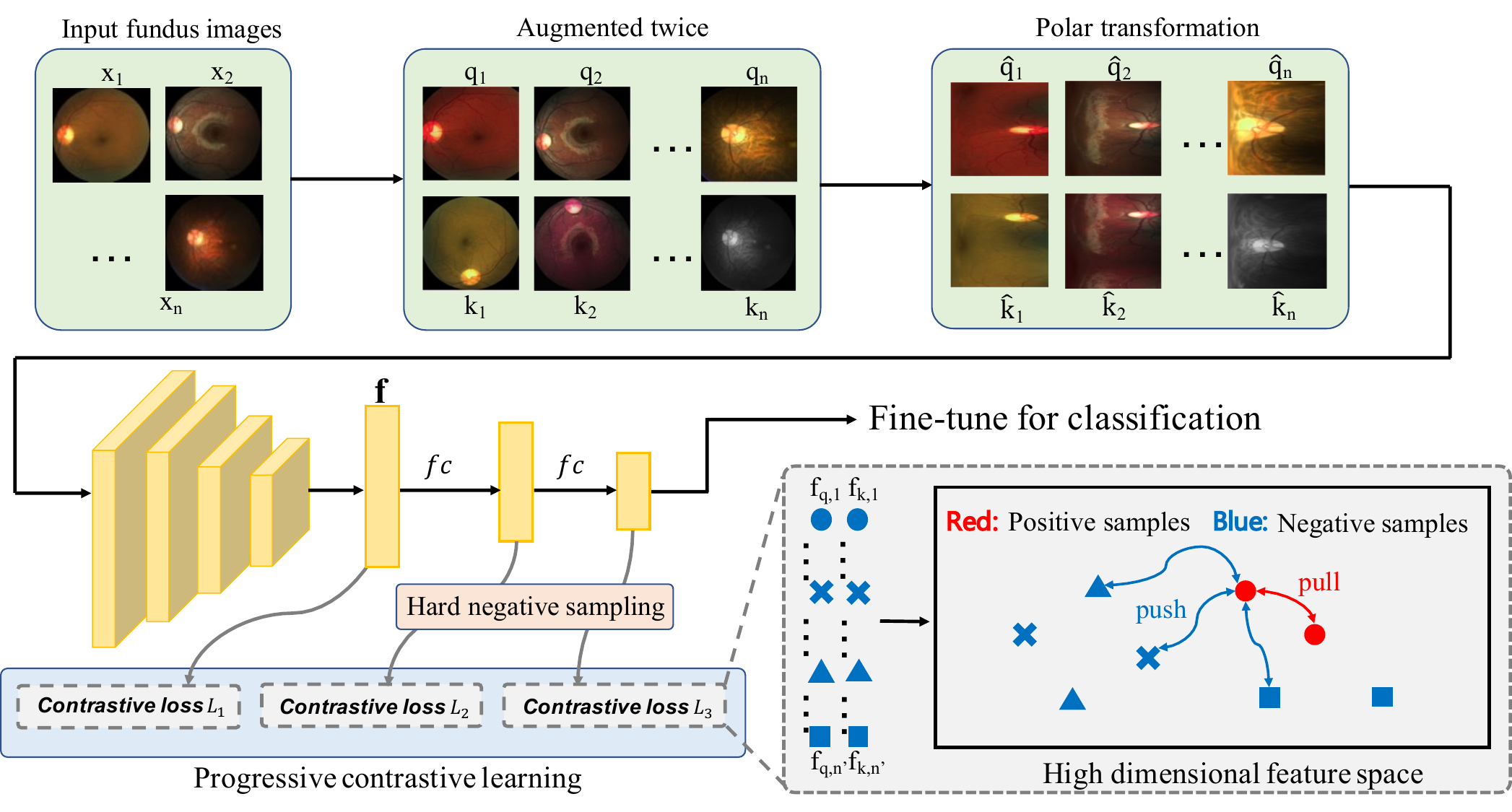}
\caption{An overview of our PoCo architecture. PoCo inputs $n$ raw fundus images in one mini-batch, and performs random data augmentation twice on each of these images to generate 
positive pairs. Then polar transformation is applied to the augmented views of the images to generate 
transformed images. The transformed images are fed to a backbone CNN to extract high-dimensional feature vectors, 
which are used for the first contrastive loss calculation. Via two FC layers, the feature vectors 
reduce their dimensions and are used to calculate the second and third contrastive losses with a hard negative sampling strategy. 
Finally, the low-dimensional features learned by PoCo are used for ophthalmic disease classification by fine-tuning the FC layers. }
\label{overview}

\end{figure*}

To address these issues, in this paper, we present a new SSL approach via \underline{Po}lar transformation-based progressive \underline{Co}ntrast learning, called PoCo, for ophthalmic disease diagnosis on fundus images. Specifically, we propose to inject polar transformation into the contrastive learning pre-training process. The polar transformation is used to transform raw fundus images to the polar coordinate system.
After this process, the rotation-invariance of the raw images is equivalent to the translation-invariance of the transformed images, while the shape of convolution scanning is equivalently transformed from square to sector. Thus, by polar transformation, rotation-invariant and annular features can be better extracted for reliable ophthalmic disease analysis. And interestingly, we find that polar transformation can promote contrastive learning pre-training to be faster and more stable. Further, we develop a progressive contrastive learning (PCL) method based on a novel progressive hard negative sampling (PHNS). PHNS removes part of negative samples and retains only some hard ones for PCL, with which the computation costs are reduced and hard negative samples are better distinguished to improve the training efficiency and performance.

Our main contributions are as follows. 
\begin{itemize}
    \item We propose a novel SSL method via polar transformation-based progressive contrastive learning for automatic ophthalmic disease diagnosis on fundus images. Our method reduces labeling effort by pre-training on unlabeled data.

    \item We propose to inject the polar transformation into contrastive learning to enhance the contrastive learning process and better extract rotation-invariant and rotation-related features for downstream tasks.

    \item We develop a progressive contrastive learning method to gradually reduce the negative sample number with a new progressive hard negative sampling for efficient training and performance improvement.

    \item We conduct extensive experiments on three public datasets to verify the superiority of our PoCo over state-of-the-art CL methods in various metrics.
\end{itemize}

\section{Related Work}
\subsection{Polar Transformation}
Polar Transformation aims to convert an image from the Cartesian coordinates system to the polar coordinates system. It has been applied in many areas for transforming feature distributions to simplify specific tasks, such as modulation classification~\cite{9188007}, and tropical cyclone analysis~\cite{chen2021cnn}. It is widely used especially in medical image analysis since the focal features of some diseases are more easily extracted after polar transformation, like brain segmentation~\cite{5762099}, disc segmentation~\cite{fu2018joint}, and glaucoma classification~\cite{lee2019screening}, etc. In ophthalmic disease, since the shape of the pupil is a circle, the polar transformation can transform some annular
into rectangular features to facilitate CNN network extraction~\cite{hu2023glim,fu2018joint}. Motivated by this, we explore utilizing polar transformation to solve a series of ophthalmic disease analysis tasks. Different from previous studies using polar transformation directly for feature extraction and classification in specific tasks, we mainly propose the application of polar transformation in contrastive learning for task-free, faster and more stable pre-training, and can better solve any downstream tasks of ophthalmic disease diagnosis.

\begin{figure*}[h]
\centering
\includegraphics[width=0.38\textwidth]{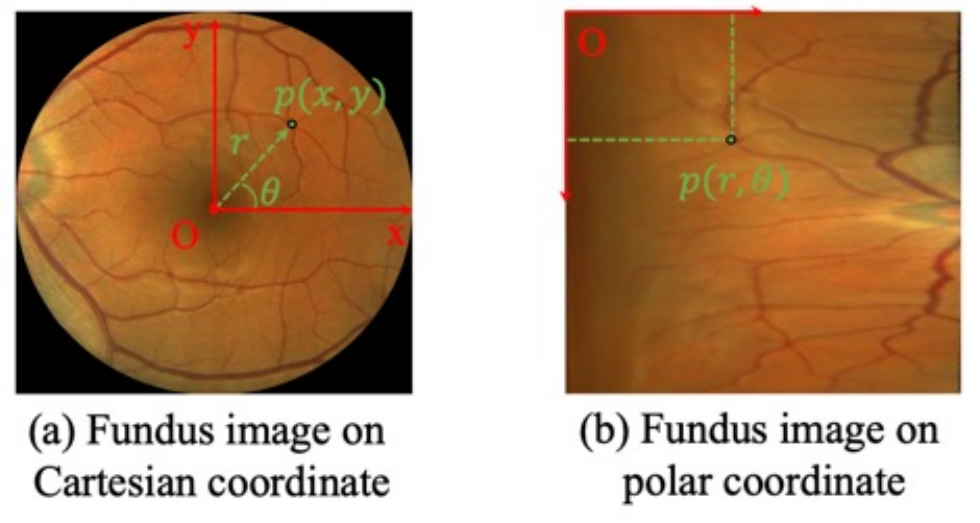}
\quad\quad
\includegraphics[width=0.38\textwidth]{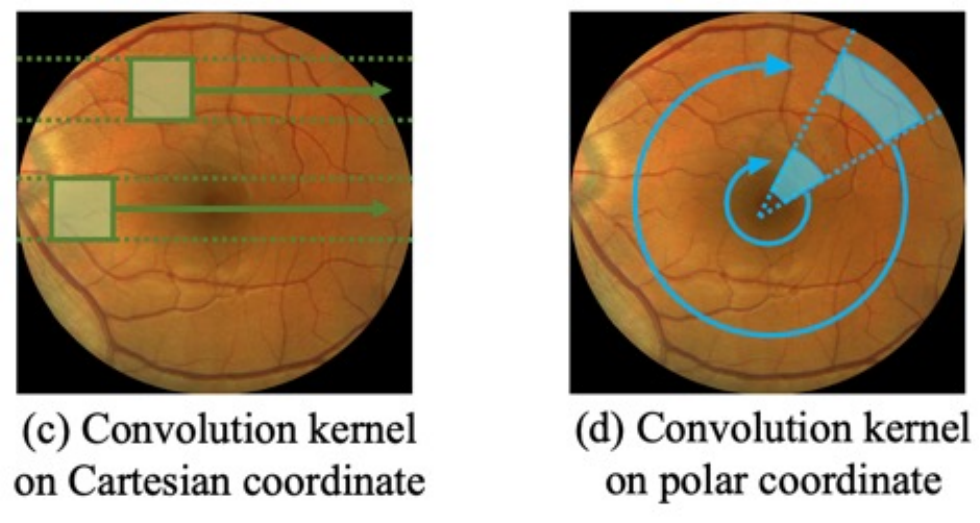}
\caption{Illustrating the pixel mapping from the Cartesian coordinate system (a) to the polar coordinate system (b) by using the polar transformation. (c)-(d) A convolution kernel working on Cartesian coordinates and polar coordinates, respectively.}
\label{polar}

\end{figure*}

\subsection{Contrastive Learning}
Contrastive learning (CL), as a type of self-supervised learning, has demonstrated the potential of similarity learning frameworks for both representation learning and downstream tasks. The goal of CL is to maximize (minimize) similarities of positive (negative) pairs at the instance level. The positive pair is only built by two correlated views of the same instance in general and the other data pairs are negative. A popular loss function is InfoNCE loss~\cite{oord2018representation}, which can pull together two data augmentation views from the same example and push away the other negative examples. MoCo~\cite{he2020momentum} proposes a memory queue to store the consistent representations. SimCLR~\cite{chen2020simple} optimizes InfoNCE within a mini-batch and has found some effective training tricks, e.g., data augmentation. 

In the ophthalmic disease diagnosis domain,  Li et al. ~\cite{li2020self} applied CL to multi-modal fundus images for retinal disease diagnosis. Li et al.~\cite{li2021rotation} devised a rotation-oriented collaborative method including a rotation prediction task and a multi-instance discrimination CL task for retinal disease diagnosis. But, these methods did not consider the specificity of ophthalmic disease analysis on fundus images. For example, the features used for diagnosis are usually annular (e.g., vessel structures), which are different from natural images. How to utilize these characteristics to enhance the effectiveness of CL is worth exploring. On the other hand, known methods also overlook that the negative hard samples could hinder the efficiency of CL pre-training and degrade the fine-tuning performance.

\section{Method}
Fig.~\ref{overview} gives an overview of our proposed PoCo for ophthalmic disease diagnosis on fundus images. {Like most CL methods, our main work aims to learn representations by maximizing agreement between differently augmented views of the same data samples via contrastive losses in
the latent space. First, in each mini-batch, we sample $n$ fundus images from the training dataset, $S=\{x_i\}_{i=1}^{n}$, and for each image $x_{i}$, apply random data augmentation (for the augmentation operation types, see the paragraph of Implementation Details in Other Details) twice to generate two images $q_{i}$ and $k_{i}$, where $q_{i}$ and $k_{i}$ are different and considered as a positive pair. Then, we apply a polar transformation to $q_{i}$ and $k_{i}$, converting Cartesian coordinates to polar coordinates. Each transformed image is fed to a CNN network to obtain a high-dimensional feature vector $\mathbf{f}$. Then, during the process of the feature vector $\mathbf{f}$ being further compressed into a lower dimensional space by fully connected (FC) layers, we perform progressive contrastive learning with a novel progressive negative sampling strategy for efficient self-supervised training. The feature $\mathbf{f}$ is decoupled to $\mathbf{f}_{q}$ and $\mathbf{f}_{k}$ for contrastive loss calculations. After that, the low-dimensional features learned by PoCo are used for ophthalmic disease diagnosis by fine-tuning the FC layers. Below we will elaborate on the polar transformation, progressive contrastive learning, and other details of our model.

\subsection{Polar Transformation (PoT)}
To better capture rotation-invariant representations for ophthalmic disease diagnosis, we propose to apply a pixel-wise polar transformation that transforms raw fundus images to the polar coordinate system.  As shown in Fig.~\ref{polar}(a), let $p(x,y)$ denote the Cartesian coordinates of a pixel $p$ in a raw fundus image, and the center pixel $o(x_0, y_0)$ of the image be at the origin of the Cartesian coordinate system. The corresponding pixel of $p$ in the polar coordinate system is $p^{\prime}(r, \theta)$ (see Fig.~\ref{polar}(b)), where
$r$ and $\theta$ are the radius and directional angle of the pixel $p$ in the raw image, respectively. We formulate the transformation relation between the Cartesian coordinates and polar coordinates as follows:
\begin{equation}
\centering
    \left\{\begin{array}{l}x=r \cos \theta \\ y=r \sin \theta\end{array} \Leftrightarrow\left\{\begin{array}{l}r=\sqrt{(x-x_0)^{2}+(y-y_0)^{2}}, \\ \theta=\tan ^{-1} (y-y_0) / (x-x_0). \end{array}\right.\right.
\end{equation}

Further, in order to retain the same size as the raw images $(H, W)$, we set the sampling distance $d$ along the radius $r_{max}$ and the sampling angle $\omega$ as follows:
\begin{equation}
\centering
    d = \frac{r_{max}}{H}, \ \omega = \frac{360}{W},
\end{equation}
where $H$ and $W$ are the height and width of the raw images, respectively. For $r_{max}$, we do not refer to the default setting where $r_{max} = \sqrt{H^{2}+W^{2}}$. Instead, we set $r_{max} = \frac{W}{2}$ since we only need the retina area which is the middle circular part of a fundus image. The polar transformation provides a pixel-wise representation of a raw image in the polar coordinate system, which offers three beneﬁts. (1) Spatial constraint: As mentioned above, the polar transformation extracts only the retina area to retain key information of the raw image and remove the useless black area around the corners. (2) Equivalent augmentation: The polar transformation is helpful for the model to efficiently learn rotation-invariant features of the raw image since it also transforms the rotation-invariance to the translation-invariance. That is, rotating raw images is equivalent to the drift transformation on the images in polar coordinates. But, rotating raw images by an arbitrary angle is difficult while translation of images after the polar transformation is simple. (3) Sector-shaped convolution: It is difficult to directly perform sector convolution on raw images, while our polar transformation is a much simpler way to transform the scanning of a convolution kernel from a square to a sector, as shown in Fig.~\ref{polar}(c) and Fig.~\ref{polar}(d). That is, the normal convolution on the transformed images is equivalent to sector-shaped convolution on the raw images. For example, sector-shaped convolution can highlight the annular morphology of the main blood vessels to better capture annular features for ophthalmic disease analysis. 

\subsection{Progressive contrastive Learning (PCL)}
To better learn the latent representation of fundus images and identify hard samples, we propose PCL to perform multi-stage contrastive learning on different dimensional features with a gradually refining negative sampling scheme. As shown in Fig.~\ref{overview}, after extracting features of an image by the CNN backbone (e.g., ResNet18~\cite{he2016deep}), the obtained feature vector will be fed to the first contrastive loss calculation. Via another two FC layers, the obtained feature vector further reduces the dimensions and is then fed to the second and third contrastive loss calculation with the progressive hard negative sample mining strategy. In this progressive process, as the dimensions of the feature vector gradually decrease, the number of negative samples decreases correspondingly. The details are given below.

\noindent
{\bf Contrastive Loss.} Contrastive learning aims to find the transformation-invariant representation based on the key hypothesis that for each image $x_i$, different data-augmented views of $x_i$ should be invariant in the latent feature space. In Fig.~\ref{overview},  $\mathbf{f}_{q,i}$ and $\mathbf{f}_{k,i}$ denote the features of the augmented views $\hat{q}_{i}$ and $\hat{k}_{i}$, respectively. We expect that $\mathbf{f}_{q,i}$ should be similar to $\mathbf{f}_{k,i}$ in a high-dimensional feature space. Formally, we define the probability of each positive pair $\mathbf{f}_{q,i}$ and $\mathbf{f}_{k,i}$ being recognized as augmented from the same raw image as: 
\vspace{-3pt}
\begin{equation}
    \begin{aligned}
        &P\left(\mathbf{f}_{q,i}|\mathbf{f}_{k,i}\right)\\
        &=\frac{\exp \left(\operatorname{sim}\left(\mathbf{f}_{q,i}, \mathbf{f}_{k,i}\right) / \tau\right)}{\exp \left(\operatorname{sim}\left(\mathbf{f}_{q,i}, \mathbf{f}_{k,i}\right) / \tau\right) + \underset{j \in S^{-}}\sum \exp \left(\operatorname{sim}\left(\mathbf{f}_{q,i}, \mathbf{f}_{k,j}\right) / \tau\right)},
    \end{aligned}
\label{eq-CL}
\end{equation}
where $\tau$ is a scalar temperature parameter, 
$\operatorname{sim}(\mathbf{f}_{q,i}, \mathbf{f}_{k,i}) = \mathbf{f}_{q,i}^{T}\mathbf{f}_{k,i}/\|\mathbf{f}_{q,i}\|\|\mathbf{f}_{k,i}\|$ denotes the cosine similarity between $\mathbf{f}_{q,i}$ and $\mathbf{f}_{k,i}$, and $S_{t}^{-}$ denotes the negative sample set for the $t$-th contrastive loss. Our goal is to increase the probability of positive sample pairs' matching and decrease the probability of negative sample pairs' matching. Thus, the final objective is to minimize the sum of the negative log-likelihood over all the images within a mini-batch, where the $t$-th contrastive loss $L_{con,t}$ can be formulated as:
\vspace{-3pt}
\begin{equation}
  \begin{aligned} 
  & \mathcal{L}_{con,t}=-\sum_{i=1}^{n} \log  P\left(\mathbf{f}_{q,i}|\mathbf{f}_{k,i}\right) -\sum_{i=1}^{n} \sum_{j \in S_{t}^{-}} \log \left(1-P\left(\mathbf{f}_{q,i}|\mathbf{f}_{k,j}\right)\right).
  \end{aligned}
  \label{eq-C-loss}
\end{equation}

It can be seen that Eq.~(\ref{eq-CL}) is formally equivalent to the Softmax function, through the loss function in Eq.~(\ref{eq-C-loss}). The model is designed to push “negative pairs” apart and pull “positive pairs” together, therefore learning the similarities of paired samples and the specificity of unpaired samples. 

\begin{table}[t]
\centering
\resizebox{0.47\textwidth}{!}{%
\begin{tabular}{l|c|c|c|c|c}
\hline
Method      & AUC & Accuracy & Recall & Precision & F1-score \\ \hline
ResNet18 & 76.1 & 82.0 & 71.1 & 62.4 & 64.9 \\ \hline
SCL
& 80.5   & 84.2        & 78.8      & 65.6         &  68.6       \\ \hline
SimCLR~
& 77.3 & 83.0  & 73.5   & 63.7      & 66.2     \\ 
LBCL
& 79.3   & 83.2        & 76.0     &    64.4      & 67.5        \\ 
FundusNet
& 80.3   & 84.0        & 78.6     & 64.7         & 67.7        \\ 
SimCLR-DR
& 80.0   & 83.5        & 78.6      & 64.5         & 67.6        \\ 
\rowcolor[HTML]{C0C0C0}
\textbf{PoCo (ours)}        & \textbf{82.0}   & \textbf{85.5}        & \textbf{80.6}      & \textbf{70.4}         & \textbf{72.3}        \\ \hline
\end{tabular}%
}
\caption{Results of different methods obtained by pre-training on the Kaggle-DR dataset and fine-tuning on the Kaggle-DR dataset (\%).}
\label{tab1}
\vskip -1 em
\end{table}

\begin{table*}[]
\centering

\resizebox{\textwidth}{!}{%
\begin{tabular}{c|ccccc|ccccc}
\hline
\multirow{2}{*}{Method} &
  \multicolumn{5}{c|}{Ichallenge-AMD} &
  \multicolumn{5}{c}{Ichallenge-PM} \\ \cline{2-11} 
 &
  \multicolumn{1}{c|}{\ AUC\ \ } &
  \multicolumn{1}{c|}{Accuracy} &
  \multicolumn{1}{c|}{Precision} &
  \multicolumn{1}{c|}{\ Recall\ \ } &
  F1-score &
  \multicolumn{1}{c|}{\ AUC\ \ } &
  \multicolumn{1}{c|}{Accuracy} &
  \multicolumn{1}{c|}{Precision} &
  \multicolumn{1}{c|}{\ Recall \ \ } &
  F1-score \\ \hline
ResNet18 (baseline) &
  \multicolumn{1}{c|}{76.51} &
  \multicolumn{1}{c|}{84.16} &
  \multicolumn{1}{c|}{82.54} &
  \multicolumn{1}{c|}{76.18} &
  78.86 &    
  \multicolumn{1}{c|}{96.01} &
  \multicolumn{1}{c|}{95.45} &
  \multicolumn{1}{c|}{94.51} &
  \multicolumn{1}{c|}{97.25} &
   95.34 \\ \hline
SimCLR~\cite{chen2020simple} &
  \multicolumn{1}{c|}{77.19} &
  \multicolumn{1}{c|}{87.09} &
  \multicolumn{1}{c|}{82.98} &
  \multicolumn{1}{c|}{77.82} &
  79.27 &
  \multicolumn{1}{c|}{98.04} &
  \multicolumn{1}{c|}{97.66} &
  \multicolumn{1}{c|}{97.30} &
  \multicolumn{1}{c|}{98.04} &
   97.53 \\
Invariant~\cite{ye2019unsupervised} &
  \multicolumn{1}{c|}{81.62} &
  \multicolumn{1}{c|}{87.51} &
  \multicolumn{1}{c|}{81.92} &
  \multicolumn{1}{c|}{81.62} &
  81.35 &
  \multicolumn{1}{c|}{98.02} &
  \multicolumn{1}{c|}{97.84} &
  \multicolumn{1}{c|}{97.56} &
  \multicolumn{1}{c|}{98.02} &
   97.75 \\
Multi-modal~\cite{li2020self} &
  \multicolumn{1}{c|}{83.17} &
  \multicolumn{1}{c|}{89.37} &
  \multicolumn{1}{c|}{85.71} &
  \multicolumn{1}{c|}{83.17} &
  83.67 &
  \multicolumn{1}{c|}{98.41} &
  \multicolumn{1}{c|}{98.38} &
  \multicolumn{1}{c|}{98.31} &
  \multicolumn{1}{c|}{98.41} &
   98.33 \\
Li et al.~\cite{li2021rotation} &
  \multicolumn{1}{c|}{84.97} &
  \multicolumn{1}{c|}{90.10} &
  \multicolumn{1}{c|}{86.11} &
  \multicolumn{1}{c|}{84.97} &
  85.27 &
  \multicolumn{1}{c|}{99.12} &
  \multicolumn{1}{c|}{99.19} &
  \multicolumn{1}{c|}{{\bfseries 99.27}} &
  \multicolumn{1}{c|}{99.12} &
   99.18 \\
Uni4Eye~\cite{cai2022uni4eye} &
  \multicolumn{1}{c|}{85.85} &
  \multicolumn{1}{c|}{90.45} &
  \multicolumn{1}{c|}{86.44} &
  \multicolumn{1}{c|}{85.85} &
  86.14 &
  \multicolumn{1}{c|}{98.53} &
  \multicolumn{1}{c|}{98.24} &
  \multicolumn{1}{c|}{97.90} &
  \multicolumn{1}{c|}{98.53} &
   98.18 \\ 
   LaCL~\cite{cheng2023lesion} &
  \multicolumn{1}{c|}{86.08} &
  \multicolumn{1}{c|}{90.60} &
  \multicolumn{1}{c|}{86.52} &
  \multicolumn{1}{c|}{86.08} &
  86.33 &
  \multicolumn{1}{c|}{98.65} &
  \multicolumn{1}{c|}{98.40} &
  \multicolumn{1}{c|}{98.00} &
  \multicolumn{1}{c|}{98.69} &
   98.32 \\
\rowcolor[HTML]{C0C0C0}
{ \textbf{PoCo (ours)}} &
  \multicolumn{1}{c|}{{\bfseries 88.30}} &
  \multicolumn{1}{c|}{{\bfseries 92.25}} &
  \multicolumn{1}{c|}{{\bfseries 86.53}} &
  \multicolumn{1}{c|}{{\bfseries 87.70}} &
  {\bfseries 88.04} &
  \multicolumn{1}{c|}{{\bfseries 99.87}} &
  \multicolumn{1}{c|}{{\bfseries 99.25}} &
  \multicolumn{1}{c|}{99.23} &
  \multicolumn{1}{c|}{{\bfseries 99.87}} &
  \multicolumn{1}{c}{{\bfseries 99.27}} \\ \hline
\end{tabular}%
}
\caption{Results of different methods obtained by pre-training on the Kaggle-DR dataset and fine-tuning on the Ichallege-AMD or Ichallenge-PM dataset (\%).}

\label{tab2}
\end{table*}

\begin{table*}[]
\centering
\resizebox{11cm}{!}{%
\begin{tabular}{c|c|c|c|c|c|c|c}
\hline
\ \ Method\ \    & \ PoT\ \  & \ PCL\ \  & \ AUC\ \    & Accuracy & Precision & \ Recall\ \  & F1-score \\ \hline
Baseline &     &     & 81.62 & 87.51    & 81.92     & 81.62  & 81.35    \\ \hline
PoCo     &   \checkmark  &     & 86.39 & 90.00    & 85.65
& 86.39  & 85.58    \\ \hline
PoCo     &     & \checkmark   & 86.52 & 90.24    & 85.73     & 86.52  & 85.90    \\ \hline
{\bfseries PoCo}     & \checkmark   &  \checkmark   & {\bfseries 88.30} & {\bfseries 92.25}    & {\bfseries 86.53}     & {\bfseries 88.30}  & {\bfseries 88.04}    \\ \hline
\end{tabular}%
}
\caption{Ablation study on the Ichallege-AMD dataset (\%). PoT = Polar Transformation; PCL = Progressive contrastive Learning.} 
\label{tab3}
\vskip -1em
\end{table*}

\noindent
{\bf Progressive Hard Sample Mining.} Hard sample mining aims to better distinguish the negative samples that are similar to positive samples (the ``hard'' negative samples). In our PCL process, we hypothesize that hard negative samples are gradually reduced due to some hard negative samples being preliminarily discriminated in the early stage of contrastive learning. Thus, we design a progressive hard negative sampling strategy in the later contrastive learning stage that reduces the size of the negative sample set $S^{-}$ by retaining only $n^\prime$ ($n^\prime < n$) hard samples that are still similar to the positive samples. For example, in the first contrastive loss calculation, $S^{-}$ contains $n-1$ negative samples (all the samples in a mini-batch except for the positive sample), but in the second and third contrastive loss calculation, $S^{-}$ contains only $n/2-1$ and $n/4-1$ negative samples, respectively. The criterion for the negative samples to be selected as hard samples is to have the largest cosine similarity value to the positive sample. To determine the best possible negative sample number $n^\prime$, we conduct experiments that explore the detailed influence of different $n^\prime$ values for the progressive contrastive learning performance in Paragraph 1 of Section Analysis. 

Like most known contrastive learning methods, our PoCo is a self-supervised pre-trained method that can learn a deep representation of images without any annotation, thus effectively reducing annotation effort. Moreover, our PoCo is better at mining hard samples that are indistinguishable and efficiently reduces computation costs by gradually reducing the number of negative samples.

\subsection{Other Details}
\label{sec-details}

\noindent
{\bf Loss Function.} For the self-supervised pre-training stage, the total objective is the sum of the three progressive contrastive losses, defined as:
\begin{equation}
   \mathcal{L}_{tot} = \mathcal{L}_{con,1} + \mathcal{L}_{con,2} + \mathcal{L}_{con,3}.
\end{equation}
In the fine-tuning stage, we use Cross-Entropy loss to train only the FC layers.

\noindent
{\bf Implementation Details.}
Our framework is built on PyTorch, and the experiments are conducted on an NVIDIA GTX 3090 GPU. For data augmentation, we resize and randomly crop images into patches of size 224 $\times$ 224. Following previous methods~\cite{li2021rotation,cai2022uni4eye}, we apply random horizontal flipping with a probability of 0.5 and random grayscaling with a probability of 0.2. Also, the brightness, contrastive, and saturation of images are changed with a random value chosen uniformly from [0.6, 1.4]. For the network architecture details, following the setting of the previous methods~\cite{ye2019unsupervised,li2021rotation}, we choose ResNet18~\cite{he2016deep} as the backbone of our network. The first and second FC layers reduce feature dimensions from 512 to 256 and from 256 to 128, respectively. The temperature $\tau$ in Eq.~(\ref{eq-CL}) is 0.5. The network is trained with the Adam optimizer~\cite{kingma2014adam} with weight decay = 0.0001 and an initial learning rate = 0.0001. The batch size is set as 64, and thus the number of negative samples is 63, 31, and 15 for the 1st, 2nd, and 3rd contrastive losses. 

\begin{figure*}[t]
\centering
\includegraphics[width=0.31\textwidth]{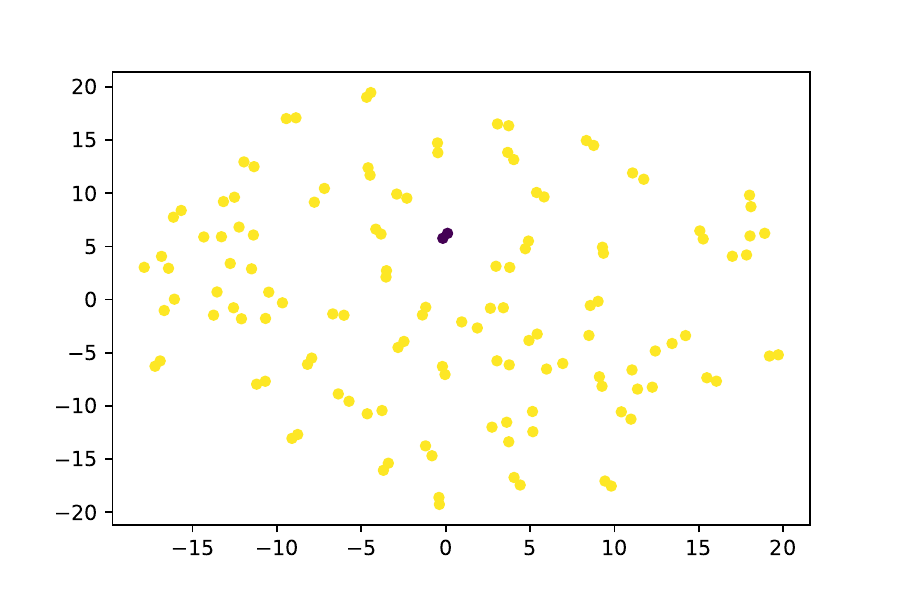}
\includegraphics[width=0.31\textwidth]{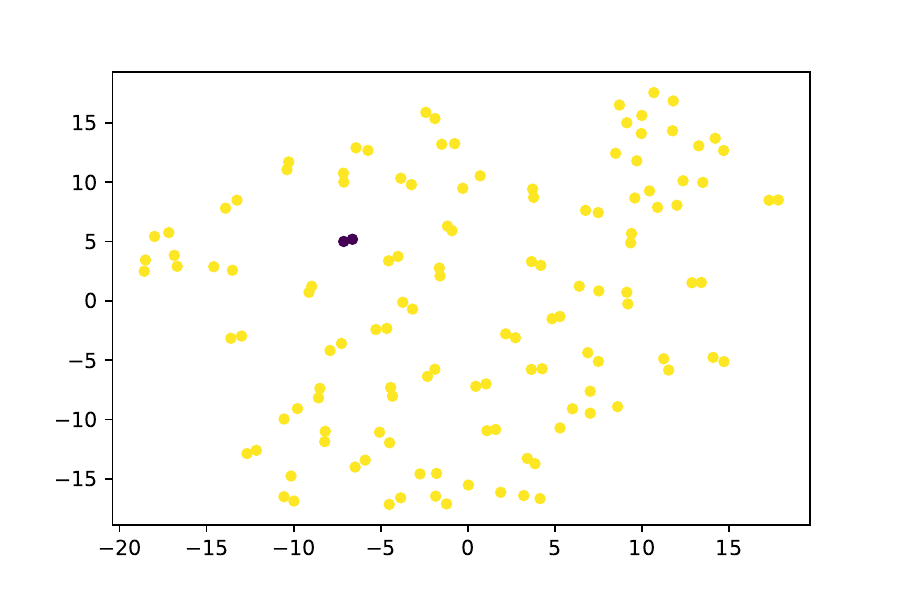}
\includegraphics[width=0.31\textwidth]{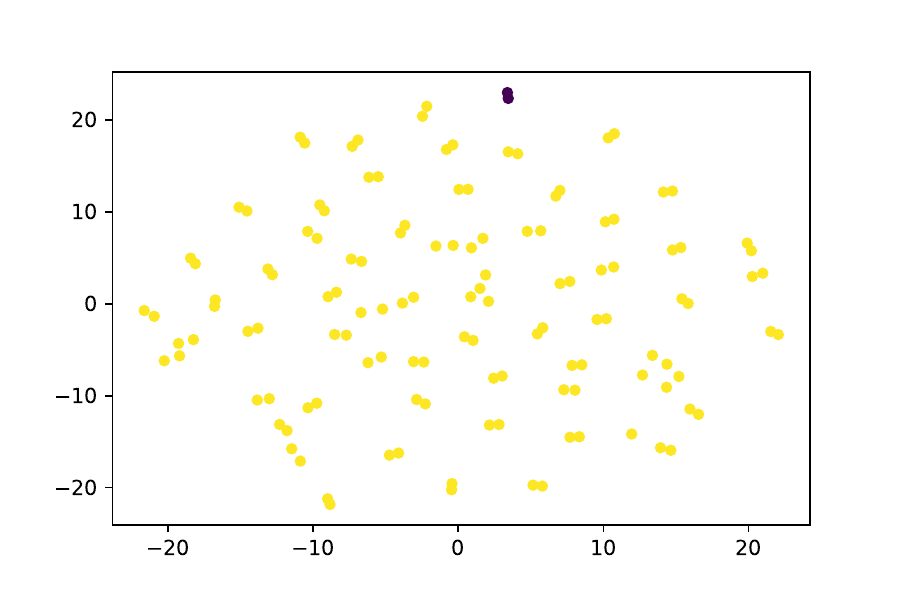}
\\ \quad (stage 1) \qquad \qquad \qquad \qquad \qquad \qquad (stage 2)  \qquad \qquad \qquad \qquad \qquad \qquad (stage 3)
\caption{Visualization of the spatial distribution of positive and negative samples at different stages by t-SNE. \textbf{\textcolor{violet}{Purple}}: Positive sample pair. \textbf{\textcolor{orange}{Yellow}}: Negative sample pairs. }
\label{vis}
\end{figure*}

\begin{table*}[t]
\centering

\resizebox{14cm}{!}{%
\begin{tabular}{ccc|c|c|c|c|c}
\hline
\multicolumn{3}{c|}{$n^\prime$ values for each stage} &
  \multirow{2}{*}{\ AUC\ \ } &
  \multirow{2}{*}{Accuracy} &
  \multirow{2}{*}{Precision} &
  \multirow{2}{*}{\ Recall\ \ } &
  \multirow{2}{*}{F1-score} \\ \cline{1-3}
\multicolumn{1}{c|}{\ \ 1st Stage\quad\quad} & \multicolumn{1}{c|}{\ \ 2nd Stage\quad\quad} & \ \ 3rd Stage\quad\quad &       &       &       &       &       \\ \hline
\multicolumn{1}{c|}{63}        & \multicolumn{1}{c|}{-}         & -         & 86.39 & 90.00    & 85.65
& 86.39  & 85.58  \\ \hline
\multicolumn{1}{c|}{63}        & \multicolumn{1}{c|}{63}        & 63        & 87.25 & 90.50 & 86.30 & 87.25 & 86.47 \\ \hline
\multicolumn{1}{c|}{{\bfseries 63}}        & \multicolumn{1}{c|}{{\bfseries 31}}        & {\bfseries 15}        & {\bfseries 88.30} & {\bfseries 92.25} & {\bfseries 86.53} & {\bfseries 88.30} & {\bfseries 88.04} \\ \hline
\multicolumn{1}{c|}{63}        & \multicolumn{1}{c|}{15}        & 3         & 87.70 & 91.50 & 86.48 & 87.70 & 86.94 \\ \hline
\multicolumn{1}{c|}{63}        & \multicolumn{1}{c|}{7}         & 1         & 87.43 & 90.50 & 86.32 & 87.43 & 86.65 \\ \hline
\end{tabular}
}
\caption{The performance of PoCo with different values of the negative sample number $n^\prime$ for each stage on the Ichallenge-AMD dataset (\%).}
\label{tab4}
\end{table*}

\section{Experiments}

\noindent
{\bf Datasets.}
We evaluate the performance of our PoCo approach using three public ophthalmic disease datasets: Kaggle-DR\footnote{\url{https://www.kaggle.com/c/diabetic-retinopathy-detection}}, Ichallenge-AMD\footnote{\url{http://ai.baidu.com/broad/introduction?dataset=amd}}~\cite{fu2020adam} and Ichallenge-PM\footnote{\url{http://ai.baidu.com/broad/introduction?dataset=pm}}~\cite{huazhu2019palm}. The details of these datasets are given as follows.
\begin{itemize}
    \item Kaggle-DR: The Kaggle-DR dataset is used for diabetic retinopathy grading, which contains 35,126 high-resolution fundus images. In this dataset, images were annotated in five levels of diabetic retinopathy from 1 to 5, representing no DR (25,810 images), mild DR (2,443 images), moderate DR (5,292 images), severe DR (873 images), and proliferative DR (708 images), respectively. This is a five-class classification task to grade the DR level that a patient has. When pre-training, all samples are used without annotations. When fine-tuning, we spilt the dataset into 6:2:2 for training, validation and test sets. 

    \item Ichallenge-AMD: The Ichallenge-AMD dataset is used for age-related macular degeneration (AMD) detection (binary classification task), which contains 1200 annotated retinal fundus images from both non-AMD subjects (77\%) and AMD patients (23\%). The training, validation, and test sets each have 400 fundus images. 

    \item Ichallenge-PM: The Ichallenge-PM dataset is used for pathological myopia (PM) detection (binary classification task), which contains 1200 annotated color fundus images with labels, including both PM and non-PM cases. All the photos were captured with Zeiss Visucam 500. The training, validation, and test sets each have 400 fundus images. 
\end{itemize}

We first train our PoCo on the Kaggle-DR dataset without annotated labels, then fine-tune PoCo on all three datasets. More details of these datasets and the experimental settings are given in the Supplementary Material.

\noindent
{\bf Metrics.} We evaluate classification performance using the metrics of Accuracy, Precision, Recall, F1-score, and AUC. AUC stands for the area under the receiver operating characteristic (ROC) curve, which measures the entire two-dimensional area underneath the entire ROC curve. ROC curve is a graphical plot that shows the diagnostic capacity of a binary classifier.

\noindent
{\bf Performance on Kaggle-DR dataset.} 
We first evaluate the performance of our PoCo on the Kaggle-DR dataset and make comparisons with other self-supervised contrastive learning methods for diabetic retinopathy grading. For fair comparison, all the models are first pre-trained on the Kaggle-DR dataset, and then fine-tuned on the Kaggle-DR dataset. ResNet18 denotes the supervised ResNet18 model which is the baseline model. SCL~\cite{feng2022robust} proposes supervised contrastive learning for diabetic retinopathy grading. Unlike our self-supervised contrastive learning, SCL utilizes ground-truths during contrastive learning pretraining. SimCLR~\cite{chen2020simple} is the classic contrastive learning method and is the contrastive learning baseline. Huang et.al~\cite{huang2021lesion} proposes a lesion-based contrastive Learning (LBCL) for
diabetic retinopathy grading. FundusNet~\cite{alam2023contrastive} applies neural style transfer to improve the performance of contrastive learning for diabetic retinopathy grading. SimCLR-DR~\cite{ouyang2023contrastive} simply applies SimCLR with knowledge transfer learning for diabetic retinopathy early detection.
The comparison results are in Table~\ref{tab1}. It can be observed that the SCL outperforms the baseline (ResNet18) model obviously since the supervised contrastive learning pre-training applied in SCL makes full use of the label's category information, but still depends on label annotations. Compared to SCL, although the self-supervised contrastive learning methods (including SimCLR, LBCL, FundusNet and SimCLR-DR) perform not as well as SCL, they still exceed the baseline model, and their pre-training process does not require annotations, which effectively saves the annotation cost. Compared to the state-of-the-art (SOTA) self-supervised contrastive learning methods FundusNet, our PoCo achieves great performance improvements of 1.7\% in AUC, 1.5\% in Accuracy, 2\% in Recall, 5.7\% in Precision and 4.6\% in F1-score, which validates the effectiveness of our proposed PoCo. Moreover, our PoCo outperforms the supervised method SCL by 1.3\% to 4.8\% in various metrics, which further demonstrates the superiority of PoCo.

\begin{figure*}[t]
\centering
\includegraphics[width=0.98\textwidth]{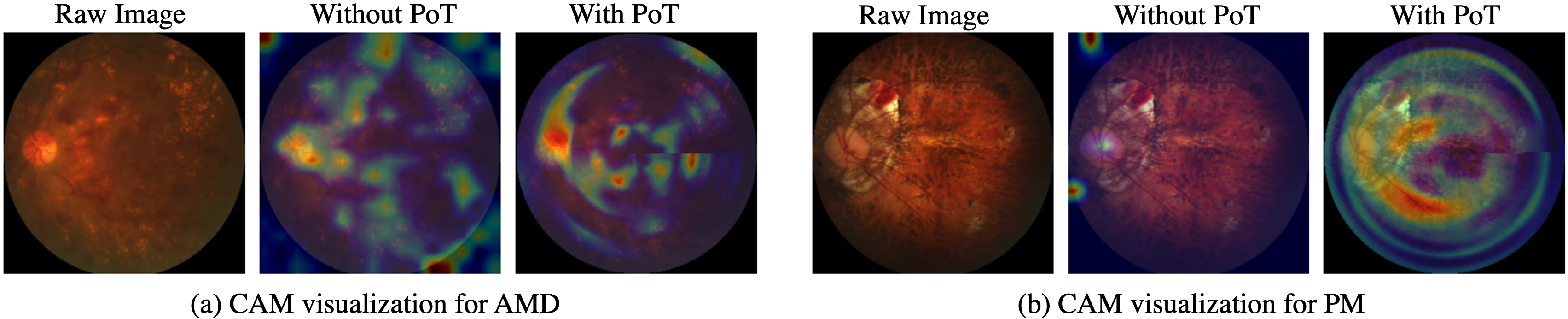}

\caption{Visualization of CAM examples of AMD (a) and PM (b) images. }
\label{cam}
\vskip -1em
\end{figure*}

\begin{figure}[t]
\centering
\includegraphics[width=0.42\textwidth]{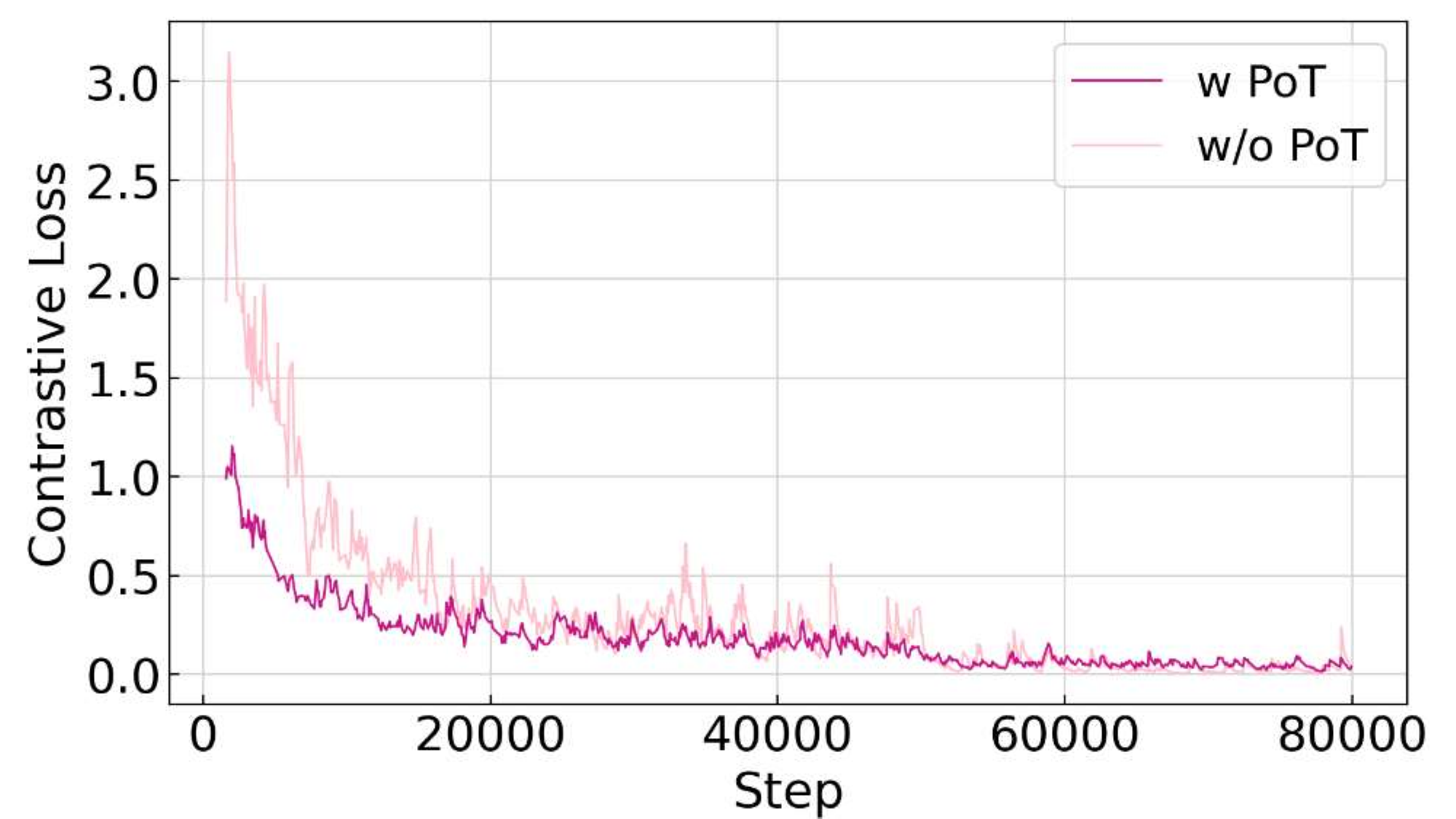}
\caption{The contrastive loss learning curves of the PoCo with and without the PoT.}
\label{loss}
\vskip -1em
\end{figure}

\noindent
{\bf Performance on Fine-tuning Generalization.} To validate the universal diagnosis capability and transfer generalization of our PoCo, we compare our PoCo with state-of-the-art (SOTA) self-supervised contrastive learning methods on fine-tuning performance on the other two ophthalmic disease datasets. For fair comparisons, all the models are first pre-trained on the Kaggle-DR dataset, and then fine-tuned on the Ichallenge-AMD and Ichallenge-PM datasets. ResNet18 denotes the supervised ResNet18 model. As shown in Table~\ref{tab2}, SimCLR~\cite{chen2020simple} and Invariant~\cite{ye2019unsupervised} yield limited performance. This is probably because the employed heavy data augmentations may not be very suitable for fundus images. The multi-modal method~\cite{li2020self} introduces additional modal information for self-supervision and gains much improvement. Li et al.~\cite{li2021rotation} obtained better results since they attempted to learn rotation-related features of fundus images. Uni4Eye~\cite{cai2022uni4eye} unifies 2D and 3D images for self-supervision and achieves SOTA performance. Although these known contrastive learning methods are beneﬁcial in improving classiﬁcation performance, it is observed that our proposed PoCo outperforms these SOTA methods and the supervised ResNet18 model, attaining improvements of 2.45\% in AUC, 1.8\% in Accuracy, and 1.9\% in F1-score on the Ichallenge-AMD dataset. On the Ichallenge-PM dataset, our method outperforms the SOTA methods in AUC by 0.75\%, Accuracy by 0.06\%, and F1-score by 0.09\%. As all methods obtain very high performance on the PM classification task, it is very challenging to make even a little improvement on the PM dataset. This demonstrates the superiority of PoCo in self-supervised contrastive learning on universal ophthalmic disease diagnosis and suggests the potential of our method on reducing annotation effort and providing reliable diagnosis.

\noindent
{\bf Ablation Study.} Next, we conduct an ablation study to validate the effect of each key component in our method. We only report the results of the Ichallenge-AMD dataset in Table~\ref{tab3}. Other results are in the Supplementary Material. We observe similar tendencies on all datasets.
The baseline applies only data augmentations without polar transformation, and calculates one contrastive loss after the final FC layer. Compared to the baseline, the PoCo with the proposed polar transformation largely improves the performance by 2.49\% to 4.77\% in various metrics, validating the effectiveness of the polar transformation for rotation-related feature extraction and can enhance contrastive learning to improve downstream task performance. The PoCo version that applies only PCL also obtains improvement, showing that progressive contrastive learning with progressive hard negative sampling is beneficial to distinguishing hard samples and improves contrastive learning performance. Moreover, by applying both polar transformation and PCL, the whole PoCo version further achieves higher performance, which demonstrates the superiority of our method and the effects of its components.

\subsection{Analysis}
\label{Ana}

\noindent
{\bf Visualization of hard negative sampling hypothesis.} To validate our hypothesis that the number of negative samples gradually decreases, we visualize the spatial distribution of positive and negative samples by t-SNE at each stage, as shown in Fig.~\ref{vis}. It is obvious that positive samples are in the center of the latent space at first, and there are many similar samples (hard negative samples) nearby. After progressive contrastive learning, positive samples gradually move to the edge of the latent space, and the nearby similar samples (hard negative samples) are gradually reduced, which makes positive samples easier to distinguish.

\noindent
{\bf Do the different negative sample numbers resulting different performances?} To explore the best \textit{Hard Negative Sampling} settings, we conduct experiments on our PoCo 
with different values of the negative sample number $n^\prime$ for each stage. Table~\ref{tab4} shows the results, from which several observations can be drawn. (1) The PoCo version with multi-stage cascade contrast learning performs better than the PoCo version with only one-stage contrast learning. This demonstrates the effectiveness of our proposed cascade contrast learning. (2) The PoCo version with different negative sample numbers in the three stages outperforms the PoCo version with the same number of negative samples in the three stages, validating that the performance promotion comes mainly from our cascade sampling strategy based on hard negative sample mining. (3) Our proposed PoCo achieves the best performance with the $n^\prime$ numbers of 63, 31, and 15 for stages 1, 2, and 3, respectively. This illustrates that PoCo is sensitive to the numbers of negative samples in the three stages, and prefers a moderate rate of negative sample number decrease. The results on the other two datasets are reported in Supplementary Material.

\noindent
{\bf Does the PoCo capture the annular features?} To further explain whether PoCo captures the annular features for disease analysis, we visualize some class activation mapping (CAM) examples of our PoCo with and without polar transformation. As shown in Fig.~\ref{cam}, the model without polar transformation extracts the discrete block area of lesions but the capture of the annular lesions is not very good. By applying the polar transformation, PoCo can better capture annular features (e.g., optic disc and blood vessels) that help accurate ophthalmic disease diagnosis, demonstrating the effectiveness of polar transformation-based contrastive learning in rotation-related feature extraction.

\noindent 
{\bf How does the Polar Transformation affect the model’s performance?} To further show the detail of why polar transformation can improve the model performance, we plot the learning curves of the total contrastive loss of the PoCo with and without polar transformation in Fig.~\ref{loss}. It can be seen that the PoCo with polar transformation has a more stable learning curve and achieves faster convergence, which we think is because polar transformation can highlight important features to reduce the fitting difficulty of contrast learning. Since the polar transformation can promote the contrastive learning of the model to be more stable and faster, the fine-tuning process can benefit from it to achieve better results.

\section{Conclusions}
In this paper, we proposed a novel SSL framework, called PoCo, for ophthalmic disease diagnosis on fundus images. Our key idea is to better capture fundus textures and learn latent invariant features in a faster and more stable way by polar transformation based progressive contrastive learning. The polar transformation extracts rotation-related features and helps the contrastive learning process to be more efficient. Further, progressive contrastive learning helps efficiently explore the transformation-invariance of different fundus images in the latent feature space by a progressive hard negative sampling strategy. Extensive experiments validated that our PoCo achieves state-of-the-art self-supervised performance and showed the potential of our method on reducing annotation effort and providing reliable diagnosis.

In the near future, we will expand our approach to other retinal diseases to show the universal diagnosis capability of PoCo on most ophthalmic diseases. Moreover, we will not only apply our method for classification but also other different tasks including object detection and segmentation.

\bibliographystyle{named}
\bibliography{ijcai24}

\begin{thebibliography}{}

\bibitem[\protect\citeauthoryear{Alakuijala \bgroup \em et al.\egroup }{1992}]{5762099}
J.~Alakuijala, J.~Oikarinen, Y.~Louhisalmi, X.~Ying, and J.~Koivukangas.
\newblock Image transformation from polar to cartesian coordinates simplifies the segmentation of brain images.
\newblock In {\em 1992 14th Annual International Conference of the IEEE Engineering in Medicine and Biology Society}, volume~5, pages 1918--1919, 1992.

\bibitem[\protect\citeauthoryear{Alam \bgroup \em et al.\egroup }{2023}]{alam2023contrastive}
Minhaj~Nur Alam, Rikiya Yamashita, Vignav Ramesh, Tejas Prabhune, Jennifer~I Lim, Robison Vernon~Paul Chan, Joelle Hallak, Theodore Leng, and Daniel Rubin.
\newblock Contrastive learning-based pretraining improves representation and transferability of diabetic retinopathy classification models.
\newblock {\em Scientific Reports}, 13(1):6047, 2023.

\bibitem[\protect\citeauthoryear{Bai \bgroup \em et al.\egroup }{2019}]{bai2019self}
Wenjia Bai, Chen Chen, Giacomo Tarroni, Jinming Duan, Florian Guitton, Steffen~E Petersen, Yike Guo, Paul~M Matthews, and Daniel Rueckert.
\newblock Self-supervised learning for cardiac {MR} image segmentation by anatomical position prediction.
\newblock In {\em Medical Image Computing and Computer Assisted Intervention--MICCAI 2019: 22nd International Conference, Shenzhen, China, October 13--17, 2019, Proceedings, Part II 22}, pages 541--549. Springer, 2019.

\bibitem[\protect\citeauthoryear{Cai \bgroup \em et al.\egroup }{2022}]{cai2022uni4eye}
Zhiyuan Cai, Li~Lin, Huaqing He, and Xiaoying Tang.
\newblock {Uni4Eye}: Unified {2D and 3D} self-supervised pre-training via masked image modeling {Transformer} for ophthalmic image classification.
\newblock In {\em Medical Image Computing and Computer Assisted Intervention--MICCAI 2022: 25th International Conference, Singapore, September 18--22, 2022, Proceedings, Part VIII}, pages 88--98. Springer, 2022.

\bibitem[\protect\citeauthoryear{Chen \bgroup \em et al.\egroup }{2015}]{chen2015glaucoma}
Xiangyu Chen, Yanwu Xu, Damon Wing~Kee Wong, Tien~Yin Wong, and Jiang Liu.
\newblock Glaucoma detection based on deep convolutional neural network.
\newblock In {\em 2015 37th Annual International Conference of the IEEE Engineering in Medicine and Biology Society (EMBC)}, pages 715--718. IEEE, 2015.

\bibitem[\protect\citeauthoryear{Chen \bgroup \em et al.\egroup }{2020}]{chen2020simple}
Ting Chen, Simon Kornblith, Mohammad Norouzi, and Geoffrey Hinton.
\newblock A simple framework for contrastive learning of visual representations.
\newblock In {\em International Conference on Machine Learning}, pages 1597--1607. PMLR, 2020.

\bibitem[\protect\citeauthoryear{chen2021cnn \bgroup \em et al.\egroup }{2021}]{chen2021cnn}
Boyo chen2021cnn, Buo-Fu Chen, and Chun~Min Hsiao.
\newblock Cnn profiler on polar coordinate images for tropical cyclone structure analysis.
\newblock In {\em Proceedings of the AAAI Conference on Artificial Intelligence}, volume~35, pages 991--998, 2021.

\bibitem[\protect\citeauthoryear{Cheng \bgroup \em et al.\egroup }{2023}]{cheng2023lesion}
Shuai Cheng, Qingshan Hou, Peng Cao, Jinzhu Yang, Xiaoli Liu, and Osmar~R Zaiane.
\newblock Lesion-aware contrastive learning for diabetic retinopathy diagnosis.
\newblock In {\em International Conference on Medical Image Computing and Computer-Assisted Intervention}, pages 671--681. Springer, 2023.

\bibitem[\protect\citeauthoryear{Feng \bgroup \em et al.\egroup }{2022}]{feng2022robust}
Xinxing Feng, Shuai Zhang, Long Xu, Xin Huang, and Yanyan Chen.
\newblock Robust classification model for diabetic retinopathy based on the contrastive learning method with a convolutional neural network.
\newblock {\em Applied Sciences}, 12(23):12071, 2022.

\bibitem[\protect\citeauthoryear{Fu \bgroup \em et al.\egroup }{2018}]{fu2018joint}
Huazhu Fu, Jun Cheng, Yanwu Xu, Damon Wing~Kee Wong, Jiang Liu, and Xiaochun Cao.
\newblock Joint optic disc and cup segmentation based on multi-label deep network and polar transformation.
\newblock {\em IEEE transactions on medical imaging}, 37(7):1597--1605, 2018.

\bibitem[\protect\citeauthoryear{Fu \bgroup \em et al.\egroup }{2020}]{fu2020adam}
H~Fu, F~Li, JI~Orlando, H~Bogunovic, X~Sun, J~Liao, Y~Xu, S~Zhang, and X~Zhang.
\newblock Adam: Automatic detection challenge on age-related macular degeneration.
\newblock {\em IEEE Dataport}, 2020.

\bibitem[\protect\citeauthoryear{Ghasemzadeh \bgroup \em et al.\egroup }{2020}]{9188007}
Pejman Ghasemzadeh, Subharthi Banerjee, Michael Hempel, and Hamid Sharif.
\newblock A novel deep learning and polar transformation framework for an adaptive automatic modulation classification.
\newblock {\em IEEE Transactions on Vehicular Technology}, 69(11):13243--13258, 2020.

\bibitem[\protect\citeauthoryear{Group and others}{2000}]{age2000risk}
Age-Related Eye Disease Study~Research Group et~al.
\newblock Risk factors associated with age-related macular degeneration: A case-control study in the age-related eye disease study: Age-related eye disease study report number 3.
\newblock {\em Ophthalmology}, 107(12):2224--2232, 2000.

\bibitem[\protect\citeauthoryear{He \bgroup \em et al.\egroup }{2016}]{he2016deep}
Kaiming He, Xiangyu Zhang, Shaoqing Ren, and Jian Sun.
\newblock Deep residual learning for image recognition.
\newblock In {\em Proceedings of the IEEE Conference on Computer Vision and Pattern Recognition}, pages 770--778, 2016.

\bibitem[\protect\citeauthoryear{He \bgroup \em et al.\egroup }{2020a}]{he2020cabnet}
Along He, Tao Li, Ning Li, Kai Wang, and Huazhu Fu.
\newblock {CABNet}: Category attention block for imbalanced diabetic retinopathy grading.
\newblock {\em IEEE Transactions on Medical Imaging}, 40(1):143--153, 2020.

\bibitem[\protect\citeauthoryear{He \bgroup \em et al.\egroup }{2020b}]{he2020momentum}
Kaiming He, Haoqi Fan, Yuxin Wu, Saining Xie, and Ross Girshick.
\newblock Momentum contrast for unsupervised visual representation learning.
\newblock In {\em Proceedings of the IEEE/CVF Conference on Computer Vision and Pattern Recognition}, pages 9729--9738, 2020.

\bibitem[\protect\citeauthoryear{Hu \bgroup \em et al.\egroup }{2023}]{hu2023glim}
Xiaoyan Hu, Ling-Xiao Zhang, Lin Gao, Weiwei Dai, Xiaoguang Han, Yu-Kun Lai, and Yiqiang Chen.
\newblock Glim-net: chronic glaucoma forecast transformer for irregularly sampled sequential fundus images.
\newblock {\em IEEE Transactions on Medical Imaging}, 2023.

\bibitem[\protect\citeauthoryear{Huang \bgroup \em et al.\egroup }{2021}]{huang2021lesion}
Yijin Huang, Li~Lin, Pujin Cheng, Junyan Lyu, and Xiaoying Tang.
\newblock Lesion-based contrastive learning for diabetic retinopathy grading from fundus images.
\newblock In {\em Medical Image Computing and Computer Assisted Intervention--MICCAI 2021: 24th International Conference, Strasbourg, France, September 27--October 1, 2021, Proceedings, Part II 24}, pages 113--123. Springer, 2021.

\bibitem[\protect\citeauthoryear{Huazhu \bgroup \em et al.\egroup }{2019}]{huazhu2019palm}
F~Huazhu, L~Fei, and IO~Jos{\'e}.
\newblock {PALM}: {PAthoLogic Myopia Challenge}.
\newblock {\em Comput. Vis. Med. Imaging}, 2019.

\bibitem[\protect\citeauthoryear{Kingma and Ba}{2014}]{kingma2014adam}
Diederik~P Kingma and Jimmy Ba.
\newblock Adam: A method for stochastic optimization.
\newblock {\em arXiv preprint arXiv:1412.6980}, 2014.

\bibitem[\protect\citeauthoryear{Lee \bgroup \em et al.\egroup }{2019}]{lee2019screening}
Jinho Lee, Youngwoo Kim, Jong~Hyo Kim, and Ki~Ho Park.
\newblock Screening glaucoma with red-free fundus photography using deep learning classifier and polar transformation.
\newblock {\em Journal of Glaucoma}, 28(3):258--264, 2019.

\bibitem[\protect\citeauthoryear{Li \bgroup \em et al.\egroup }{2019}]{li2019canet}
Xiaomeng Li, Xiaowei Hu, Lequan Yu, Lei Zhu, Chi-Wing Fu, and Pheng-Ann Heng.
\newblock {CANet}: Cross-disease attention network for joint diabetic retinopathy and diabetic macular edema grading.
\newblock {\em IEEE Transactions on Medical Imaging}, 39(5):1483--1493, 2019.

\bibitem[\protect\citeauthoryear{Li \bgroup \em et al.\egroup }{2020}]{li2020self}
Xiaomeng Li, Mengyu Jia, Md~Tauhidul Islam, Lequan Yu, and Lei Xing.
\newblock Self-supervised feature learning via exploiting multi-modal data for retinal disease diagnosis.
\newblock {\em IEEE Transactions on Medical Imaging}, 39(12):4023--4033, 2020.

\bibitem[\protect\citeauthoryear{Li \bgroup \em et al.\egroup }{2021}]{li2021rotation}
Xiaomeng Li, Xiaowei Hu, Xiaojuan Qi, Lequan Yu, Wei Zhao, Pheng-Ann Heng, and Lei Xing.
\newblock Rotation-oriented collaborative self-supervised learning for retinal disease diagnosis.
\newblock {\em IEEE Transactions on Medical Imaging}, 40(9):2284--2294, 2021.

\bibitem[\protect\citeauthoryear{Morgan \bgroup \em et al.\egroup }{2012}]{morgan2012myopia}
Ian~G Morgan, Kyoko Ohno-Matsui, and Seang-Mei Saw.
\newblock Myopia.
\newblock {\em The Lancet}, 379(9827):1739--1748, 2012.

\bibitem[\protect\citeauthoryear{Oord \bgroup \em et al.\egroup }{2018}]{oord2018representation}
Aaron van~den Oord, Yazhe Li, and Oriol Vinyals.
\newblock Representation learning with contrastive predictive coding.
\newblock {\em arXiv preprint arXiv:1807.03748}, 2018.

\bibitem[\protect\citeauthoryear{Ouyang \bgroup \em et al.\egroup }{2023}]{ouyang2023contrastive}
Jihong Ouyang, Dong Mao, Zeqi Guo, Siguang Liu, Dong Xu, and Wenting Wang.
\newblock Contrastive self-supervised learning for diabetic retinopathy early detection.
\newblock {\em Medical \& Biological Engineering \& Computing}, pages 1--12, 2023.

\bibitem[\protect\citeauthoryear{Sahlsten \bgroup \em et al.\egroup }{2019}]{sahlsten2019deep}
Jaakko Sahlsten, Joel Jaskari, Jyri Kivinen, Lauri Turunen, Esa Jaanio, Kustaa Hietala, and Kimmo Kaski.
\newblock Deep learning fundus image analysis for diabetic retinopathy and macular edema grading.
\newblock {\em Scientific Reports}, 9(1):10750, 2019.

\bibitem[\protect\citeauthoryear{Tajbakhsh \bgroup \em et al.\egroup }{2019}]{tajbakhsh2019surrogate}
Nima Tajbakhsh, Yufei Hu, Junli Cao, Xingjian Yan, Yi~Xiao, Yong Lu, Jianming Liang, Demetri Terzopoulos, and Xiaowei Ding.
\newblock Surrogate supervision for medical image analysis: Effective deep learning from limited quantities of labeled data.
\newblock In {\em 2019 IEEE 16th International Symposium on Biomedical Imaging (ISBI 2019)}, pages 1251--1255. IEEE, 2019.

\bibitem[\protect\citeauthoryear{Ye \bgroup \em et al.\egroup }{2019}]{ye2019unsupervised}
Mang Ye, Xu~Zhang, Pong~C Yuen, and Shih-Fu Chang.
\newblock Unsupervised embedding learning via invariant and spreading instance feature.
\newblock In {\em Proceedings of the IEEE/CVF Conference on Computer Vision and Pattern Recognition}, pages 6210--6219, 2019.

\bibitem[\protect\citeauthoryear{Yim \bgroup \em et al.\egroup }{2020}]{yim2020predicting}
Jason Yim, Reena Chopra, Terry Spitz, Jim Winkens, Annette Obika, Christopher Kelly, Harry Askham, Marko Lukic, Josef Huemer, Katrin Fasler, et~al.
\newblock Predicting conversion to wet age-related macular degeneration using deep learning.
\newblock {\em Nature Medicine}, 26(6):892--899, 2020.

\bibitem[\protect\citeauthoryear{Zhang \bgroup \em et al.\egroup }{2019}]{zhang2019automatic}
Hongyan Zhang, Kai Niu, Yanmin Xiong, Weihua Yang, ZhiQiang He, and Hongxin Song.
\newblock Automatic cataract grading methods based on deep learning.
\newblock {\em Computer Methods and Programs in Biomedicine}, 182:104978, 2019.

\bibitem[\protect\citeauthoryear{Zhuang \bgroup \em et al.\egroup }{2019}]{zhuang2019self}
Xinrui Zhuang, Yuexiang Li, Yifan Hu, Kai Ma, Yujiu Yang, and Yefeng Zheng.
\newblock Self-supervised feature learning for {3D} medical images by playing a {Rubik’s} cube.
\newblock In {\em Medical Image Computing and Computer Assisted Intervention--MICCAI 2019: 22nd International Conference, Shenzhen, China, October 13--17, 2019, Proceedings, Part IV 22}, pages 420--428. Springer, 2019.

\end{thebibliography}

\end{document}


\title{Supplemental Document for Submission \#166: \\``PoCo: A Self-Supervised Approach via Polar Transformation Based Progressive Contrastive Learning for Ophthalmic Disease Diagnosis''}

\maketitle              

\section{Dataset Details and Experimental Settings}

\subsection{Kaggle-DR Dataset}

\begin{figure}[h]
\centering
\includegraphics[width=0.95\textwidth]{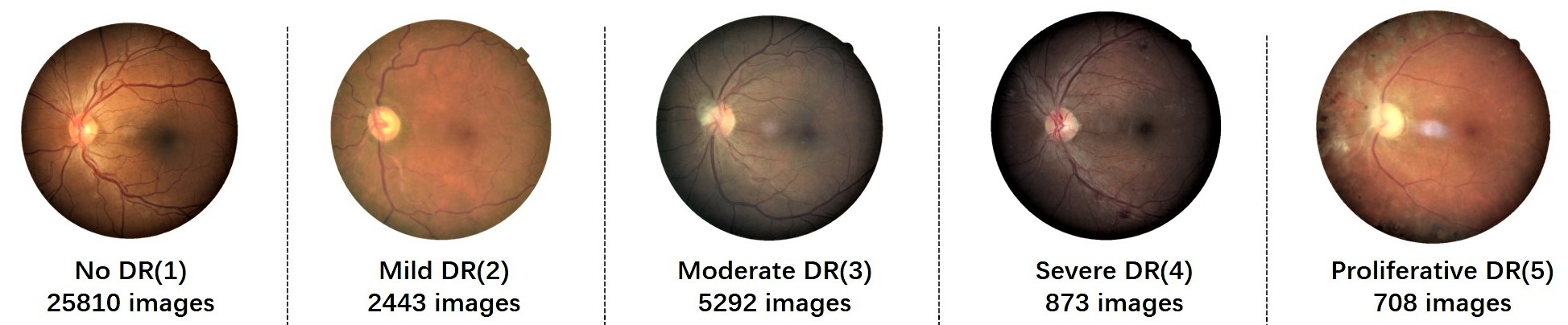}
\caption{Some sample fundus images with different diabetic retinopathy levels of the Kaggle-DR dataset.}
\label{dr}
\end{figure}

\noindent
{\bf Dataset Details.} 
The Kaggle-DR dataset contains 35,126 high-resolution fundus images. In this dataset, images were annotated in five levels of diabetic retinopathy from 1 to 5, representing no DR (25,810 images), mild DR (2,443 images), moderate DR (5,292 images), severe DR (873 images), and proliferative DR (708 images), respectively. Some examples are shown in Fig.~\ref{dr}. Note that we only use the images of this dataset without any annotations to pre-train our PoCo. Then, we subsequently fine-tune PoCo on the Ichallenge-AMD and Ichallenge-PM datasets, and report classification results on these two datasets.

\noindent
{\bf Experimental Settings.} 
For fair comparison, following the previous work, we pre-train all unsupervised models on the Kaggle-DR dataset for 150 epochs. We then freeze the model parameters and only evaluate the model performance on the target datasets by fine-tuning the fully connected (FC) layers.

\subsection{Ichallenge-AMD Dataset}

\noindent
{\bf Dataset Details.} 
The Ichallenge-AMD dataset contains 1200 annotated retinal fundus images from both non-AMD subjects (77\%) and AMD patients (23\%). The training, validation, and test sets each have 400 fundus images. Since only training data is released with annotations, we apply 5-fold cross-validation on the training set.

\noindent
{\bf Experimental Settings.} 
To fairly compare with the self-supervised methods, when fine-tuning, we train all the models for 2000 epochs. We resize the input images to 224 $\times$ 224, and apply only random horizontal flipping in data augmentation, as in previous works.

\subsection{Ichallenge-PM Dataset}

\noindent
{\bf Dataset Details.} 
The Ichallenge-PM dataset contains 1200 annotated color fundus images with labels, including both PM and non-PM cases. All the photos were captured with Zeiss Visucam 500. The training, validation, and test sets each have 400 fundus images. Since only training data is released with annotations, we apply 5-fold cross-validation on the training set.

\noindent
{\bf Experimental Settings.} 
The experimental settings are the same as the settings for the Ichallenge-AMD dataset, as described above.
